\documentclass[letterpaper, 10 pt, journal, twoside]{IEEEtran}  

\usepackage{graphicx} 
\graphicspath{{figures/}}

\usepackage{amsthm}
\usepackage{amsmath} 
\usepackage{amssymb}  
\usepackage{mathtools}
\usepackage{ulem} 
\usepackage{multirow}
\usepackage{verbatim}
\usepackage{soul}
\usepackage{algorithm}
\usepackage{algpseudocode}

\usepackage{color}
\usepackage{xcolor}
\usepackage{latexsym}
\usepackage{multicol}
\usepackage{booktabs}
\usepackage{adjustbox}

\usepackage{enumitem}
\usepackage[font=small,labelfont=bf]{caption}
\usepackage{lipsum}
\usepackage{subcaption}



\renewcommand{\thefigure}{\arabic{figure} }

\IEEEoverridecommandlockouts                              

\begin{document}

\title{\LARGE \bf
FASTNav: Fine-tuned Adaptive Small-language-models Trained for Multi-point Robot Navigation
}

\author{Yuxuan Chen$^{1*}$, Yixin Han$^{1*}$ and Xiao Li$^{1}$

\thanks{$^{*}$ denotes equal contribution.}
\thanks{$^{1}$ School of Mechanical Engineering, Shanghai Jiao Tong University
         {\tt\footnotesize \{chen\_yuxuan, hanyixin, sjtu\_lixiao\}@sjtu.edu.cn}}%

}


\maketitle
\thispagestyle{empty}
\pagestyle{empty}

\begin{abstract}
With the rapid development of large language models (LLM), robots are starting to enjoy the benefits of new interaction methods that large language models bring. Because edge computing fulfills the needs for rapid response, privacy, and network autonomy, we believe it facilitates the extensive deployment of large models for robot navigation across various industries. To enable local deployment of language models on edge devices, we adopt some model boosting methods. In this paper, we propose \textit{FASTNav} - a method for boosting lightweight LLMs, also known as small language models (SLMs), for robot navigation. The proposed method contains three modules: fine-tuning, teacher-student iteration, and language-based multi-point robot navigation. We train and evaluate models with FASTNav in both simulation and real robots, proving that we can deploy them with low cost, high accuracy and low response time. Compared to other model compression methods, FASTNav shows potential in the local deployment of language models and tends to be a promising solution for language-guided robot navigation on edge devices.
\end{abstract}

\section{INTRODUCTION}
The adoption of robots has significantly expanded across various domains, including medical, household, and industrial sectors, driven by the rapid advancements in robotics technology. However, traditional robot control methods are limited to simple tasks and are inadequate for handling complex environments and tasks described in natural language. Large language models (LLMs) such as ChatGPT and Llama have demonstrated exceptional natural language processing and logical reasoning abilities, highlighting their potential for enhancing robot navigation. Current approaches predominantly depend on API calls to leverage LLMs, which pose challenges in terms of privacy protection and real-time responsiveness, thereby limiting their practical application. In this work, we try to address the following question: ``\textit{can we develop a robot navigation system based on language models that (1) is lightweight and deployable locally, and (2) is computationally inexpensive while exhibiting high performance for complicated navigation tasks?}"

Recently, with the integration of robot control and LLMs, many innovative ideas have emerged. Some zero-shot or few-shot methods that use pre-trained models for navigation without the need for fine-tuning or annotating data \cite{LMNav2022} \cite{Saycan2022} have emerged. Additionally, there are methods that use spatial map representations to directly combine pre-trained visual-linguistic features with 3D reconstructions of the physical world \cite{VLMaps2023}. While these methods excel in perceiving the real world, they exhibit inefficiencies in the LLM component, as indicated by the ``fast-forward" signs frequently seen in demonstration videos. By reducing the size and complexity of large models, model compression methods can significantly decrease computational load and inference time, offering a promising solution to address those inefficiencies. Additionally, these techniques enable the deployment of lightweight models on edge devices, ensuring more efficient resource usage without sacrificing too much performance. However, when compressing, most of these methods require long running times and high computing power and do not perform well for specific tasks.

 \begin{figure}[t]
    \centering
    \includegraphics[width=1\linewidth]{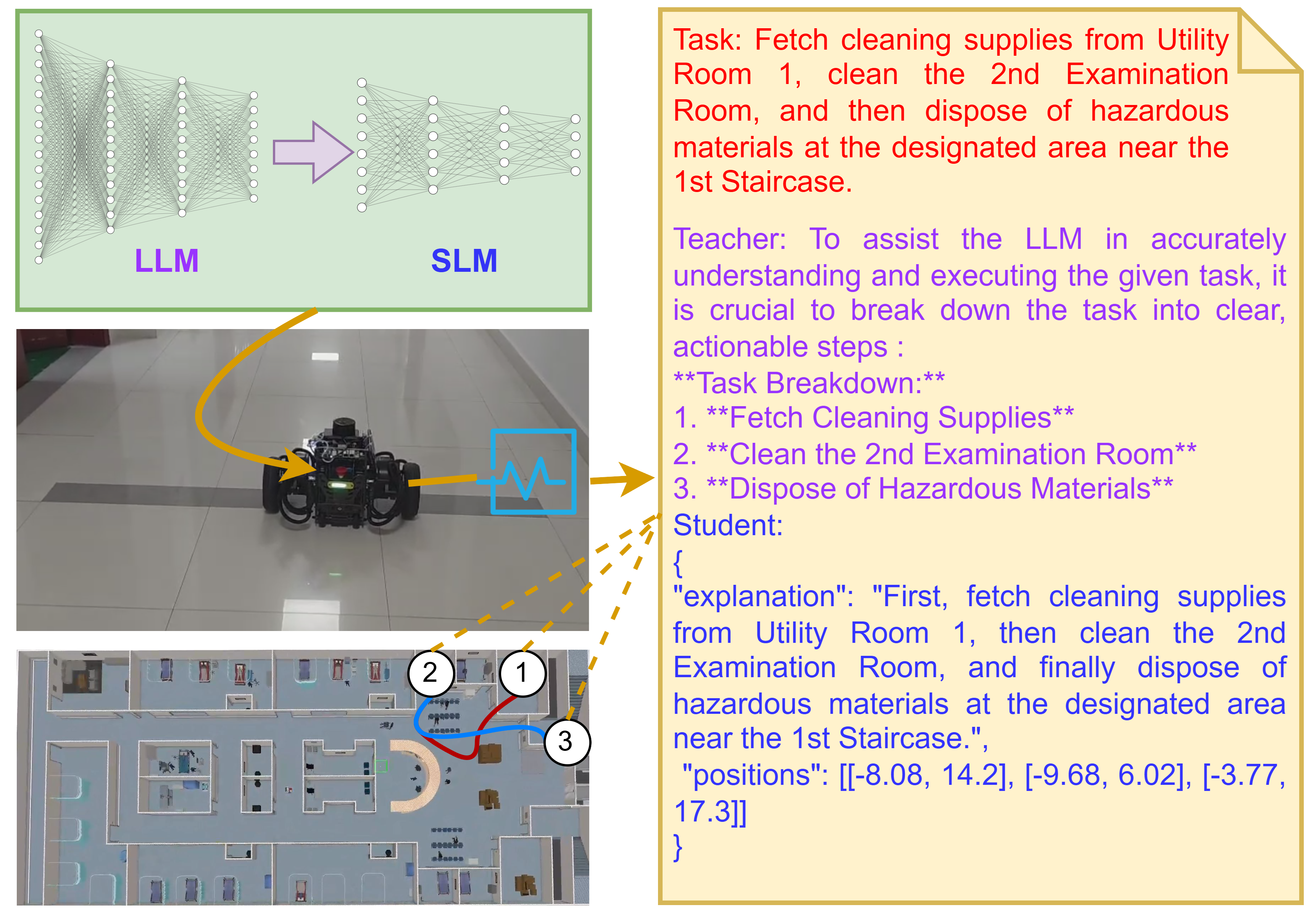}
    \caption{\textbf{FASTNav} is a method for boosting SLMs for robot navigation. We improve the performance of SLMs for multi-point robot navigation by fine-tuning and teacher-student iteration.}
    \label{fig:f1}
\end{figure}

To solve all these problems above, we propose a technology of model boosting. We draw inspiration from model compression, leveraging knowledge transfer from LLMs to small language models (SLMs) through prompting and feedback. This way enhances the capabilities of the smaller model while preserving its lightweight nature. \textbf{Our approach involves first fine-tuning the SLMs using domain-specific datasets and then applying a teacher-student iteration module to further improve their performance.} This enables SLMs to achieve performance levels close to those of much larger models in robot navigation tasks, while also offering high feasibility for practical deployment. The illustration of our system is shown in Figure \ref{fig:f1}.

To address the challenge of deploying language models locally on robots, we select smaller models and fine-tune them with LoRA \cite{Hu2021LoRALA}. This enhances the models' inference speed and accuracy without requiring extensive prompt engineering. Additionally, we design a teacher-student iteration module to enable the models to continuously learn from navigation tasks. To summarize our contributions, we
\begin{itemize}
    \item explored an innovative method of model boosting that boosts the performance of SLMs to levels comparable to those of much larger language models for specific domain tasks;
    \item introduced FASTNav, a novel method utilizing fine-tuned adaptive small-language-models (SLMs) for multi-point robot navigation;
    \item evaluated FASTNav in both simulated and real environments, demonstrating significant improvements in success rates and implementation efficiency, leading to a network-independent, large model-based robot navigation approach that ensures user privacy and data security while providing a cost-effective, high-performance solution.
\end{itemize}

\noindent In this work, we focus on boosting SLMs to bring their performance in robot navigation closer to that of much larger models, enabling robots to interact with users accurately and promptly.

\section{RELATED WORK}\label{sec:literature review}

\textbf{Using LLMs for robot navigation} has become a significant research field in robotics control technology, significantly impacting autonomous driving technologies. Target navigation points can be determined based on users' instructions using vision \cite{vision2022}, landmark \cite{landmarknav}, olfaction \cite{olfactionnav} and large language model technologies \cite{lmnav2023} \cite{wang2024srlm}. In recent years, technologies combining LLMs with visual modules have been widely used in robot navigation \cite{Hao_2020_CVPR} \cite{LMNav2022}, showing excellent results. \textit{However, constrained by the parameter sizes of small models, these methods do not show significant improvements on local small models.} 

\textbf{Applying LLMs to planning tasks} represents a growing area of research aimed at enhancing the capabilities of autonomous agents. Some methods leverage LLMs to translate planning problems into Planning Domain Definition Language (PDDL) format, achieving strong performance on specific tasks \cite{LLM+P} \cite{Dagan2023DynamicPW}. Some other approaches use Signal Temporal Logic (STL) as an intermediate representation, endowing LLMs with much stronger reasoning and planning capabilities than natural language alone \cite{AutoTAMP}. \textit{Even so, these methods face challenges such as high computational costs, slow response times, and limited adaptability in complex environments.}

\textbf{Model compression methods for LLMs} are already numerous. Pruning reduces the size and computational cost of a neural network by removing less important weights or neurons while attempting to preserve the model's performance \cite{PruningML} \cite{LLMPrunerOT}. Knowledge distillation transfers the knowledge from a large model (teacher model) to a smaller model (student model), enabling the smaller model to achieve comparable performance \cite{KnowledgeDO} \cite{IncontextLD}. Quantization compresses a model by reducing the precision of its weights and activations, typically by converting them from floating-point to lower-bit representations, thereby saving memory and computational resources \cite{Frantar2022GPTQAP} \cite{LLMQATDQ}. Low-rank factorization approximates the weight matrices of a model with lower-rank matrices, reducing the number of parameters and computational complexity without significantly degrading performance \cite{ZeroQuantFPAL}. \textit{However, most of these methods require long running times and high computing power requirements, and do not perform well for specific tasks.}

\textbf{Navigation with SLMs }can speed up operation and benefit privacy protection. Research on controlling robots with local SLMs is rare, and methods to improve the performance of local small models are scarce. To fill this gap, we start to attempt related research. We aim to enhance the performance of local models through various comprehensive methods, including fine-tuning language models using datasets, supporting them with prompt engineering, and exploring the limits of local SLMs' performance using the teacher-student iteration, to improve accuracy in multi-target navigation tasks.

\begin{figure*}[htb]
    \centering
    \includegraphics[width=1.9\columnwidth]{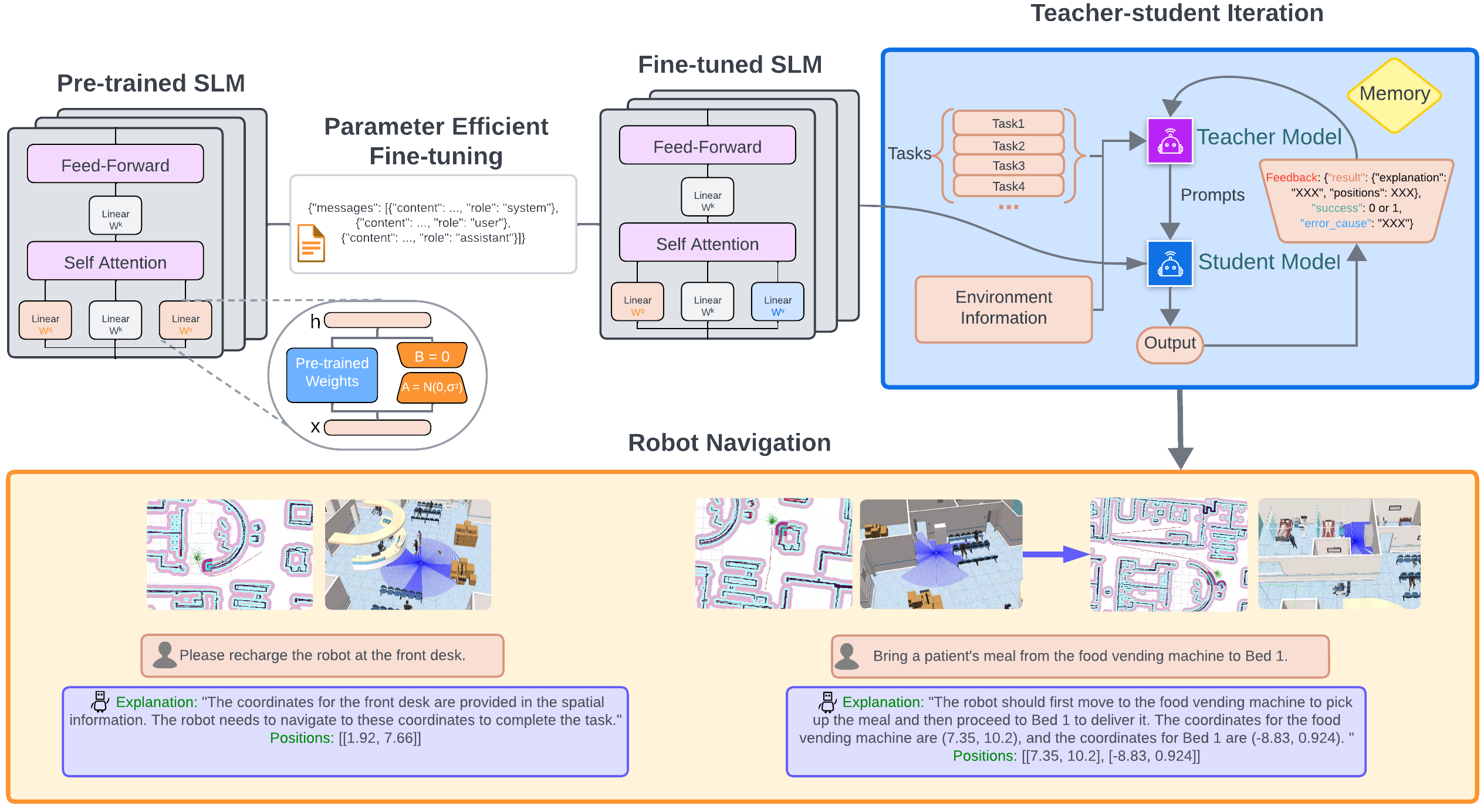}
    \caption{\textbf{Architecture of FASTNav.} The architecture contains three components:  (1) fine-tuning of SLMs based on PEFT: we use PEFT to give SLMs access to the necessary environment information and to constrain their output format; (2) teacher-student iteration: the fine-tuned SLMs cooperate with teacher models to solve navigation tasks, where the teacher serves as a prompt engineer and feedback receiver, transferring knowledge and helping SLMs correct mistakes and (3) robot navigation: the robot receives an ordered list of coordinate points from SLMs and moves with the navigation algorithms according to the list.}
    \label{fig:architecture}
\end{figure*}

\section{BACKGROUND}
\label{sec:background}

\subsection{Parameter-Efficient Fine-Tuning Based on LoRA}
\label{sec:fine-tuning}

A key issue with SLMs is their limited logical reasoning and analytical capabilities. Additionally, their output format may not be robust enough for direct application to domain-specific tasks. Therefore, it is important to fine-tune SLMs to adapt them to the specific domain of robot navigation. Some researches \cite{Mitra2023Orca2T} \cite{Juneja2023SmallLM} have shown that with appropriate fine-tuning, SLMs can achieve performance close to that of LLMs.

The fine-tuning method we use is LoRA\cite{Hu2021LoRALA}, Low-Rank Adaptation of LLMs, which is an innovative method for parameter-efficient fine-tuning.

LoRA represents parameter updates $\Delta \Phi$ with a much smaller set of parameters compared to the full set $\Theta$. This is done by using a low-rank matrix to encode the weight update $\Delta W$ as:

\begin{equation}
W_0 + \Delta W = W_0 + BA
\end{equation}

where $W_0$ is the pre-trained weight matrix, $B \in \mathbb{R}^{d \times r}$, and $A \in \mathbb{R}^{r \times k}$, with $r \ll \min(d, k)$. During training, $W_0$ is kept frozen, and only $A$ and $B$ are updated.

\subsection{Language Navigation for Robots}
We conceptualize the Language Navigation for Robot problem as follows: Given a natural language command $W$, comprised of a sequence of words $w_1, w_2, w_3, \ldots, w_{n_w}$, the language model is tasked with making navigation decisions based on a pre-constructed map representation $M$. Upon receiving the command, the model leverages $M$ to determine a series of target waypoints that define the navigation path.

The decision-making process can be formalized by the function $f$, which maps the language command and the map information to the sequence of waypoints: $f(W, M) = \{p_1, p_2, \ldots, p_n\}$, where each $p_i$ is a coordinate or identifier for a location on the map $M$.

The primary objective of the language model is to optimize the sequence of waypoints such that it accurately reflects the intent of the language directive while adhering to the constraints and affordances of the map $M$. Therefore, the most critical formula representing the decision-making function in isolation is:

\begin{equation}
f(W, M) = \{p_1, p_2, \ldots, p_n\}
\end{equation}

This function encapsulates the essence of this problem, translating a linguistic input into a navigational output within the context of a given spatial representation.

\section{PROBLEM DEFINITION AND APPROACH}
\label{sec:problem}

Our goal is to fetch a lightweight SLM that performs well for robot navigation tasks by fine-tuning and teacher-student iteration. Formally, the input is a natural language task description \(L\) and map information \(M\), and the output is a sequence of target coordinates \(T = \{t_1, t_2, \dots, t_k\}\) arranged in order.

First, we fine-tune the SLM $f_{\theta_s}(L, M)$, where $L$ is the task description and $M$ is the map data. The goal here is to minimize the fine-tuning loss $\mathcal{L}_{\text{FT}}$, ensuring the model can accurately predict the sequence of target coordinates $T = \{t_1, t_2, \dots, t_k\}$ for navigation.

Once fine-tuning is complete, we improve the SLM by learning from a larger model $f_{\theta_l}(L, M)$. This process tries to reduce both the task-specific loss $\mathcal{L}_{\text{FT}}$ and the knowledge distillation loss $\mathcal{L}_{\text{TS}}$, allowing the small model to refine its predictions based on the larger model's output.

Our approach ensures the small model can effectively generate robot navigation paths from natural language descriptions and map data. It is assumed that most objects' positions on the map (including landmarks) remain stable, though it allows for the possibility of small, dynamic obstacles. By combining fine-tuning with knowledge from the larger model, we aim to enhance the small model’s performance while maintaining its efficiency for real-world tasks.

\section{FASTNav}

In this section, we present the general structure of our approach. FASTNav is divided into three main sections: (1) fine-tuning of small language models; (2) teacher-student iteration and (3) robot navigation controller. Figure \ref{fig:architecture} provides a general overview of the structure.

\subsection{Fine-tuning of SLMs}
\label{sec:fine-tuning of SLMs}

\begin{figure}[htb]
    \centering
    \includegraphics[width=0.8\linewidth]{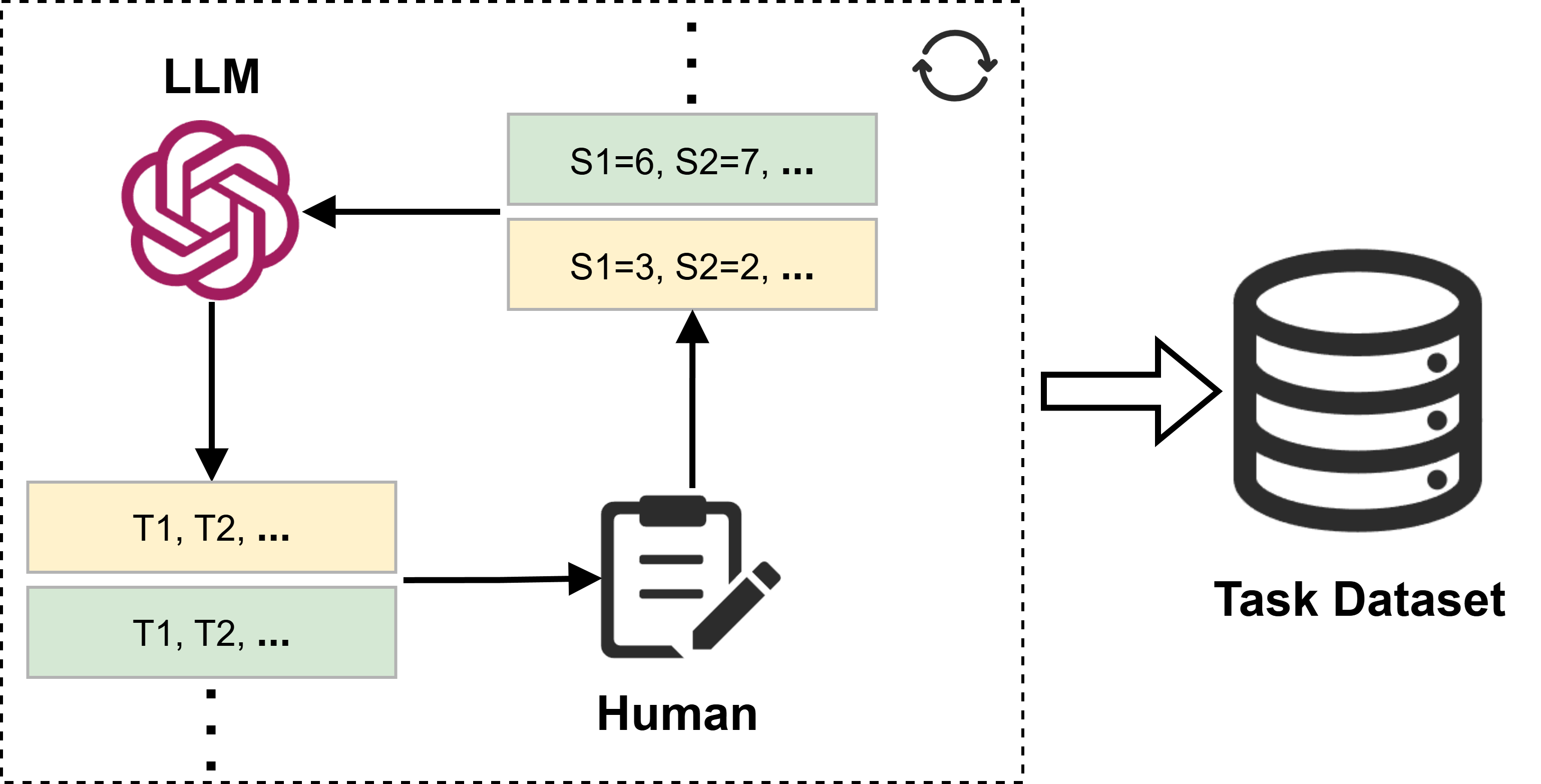}
    \caption{\textbf{Human-in-the-loop generation of the dataset.} The human-in-the-loop generation is used to generate our dataset for fine-tuning, which is a valid method for obtaining datasets aligned with humans. The yellow box corresponds to the first cycle, and the green box corresponds to the second cycle. $T_i$ means the i-th task while $S_i$ means the score that $T_i$ gets.}
    \label{fig:dataset}
\end{figure}

Considering the shortcomings that SLMs have when dealing with specific tasks, a fine-tuning process is needed to adapt them to the specific domain of robot navigation.

During the fine-tuning process, the expected output of language models is a JSON format context that includes \textit{explanation} and \textit{positions}. The \textit{explanation} is a string that shows the model's analysis and reasoning about the task, making it easy to understand its thought process and to see why the task succeeds or fails. The \textit{positions} is a list of pairs of numbers containing the $x$ and $y$ coordinates of the target points. We hope SLMs can output in such a format, utilizing the \textit{positions} to navigate the robot and showing its thought process through the \textit{explanation}. 

Human-in-the-loop generation is used to generate our dataset for fine-tuning. We first write some tasks using examples from real life. The tasks and some information about the environment are composed to make a prompt. Then an LLM generates some tasks according to the prompt. A human evaluator joins the process and gives scores to the tasks. The tasks and the scores are brought back to the LLM, and it generates a set of new tasks with this feedback from humans. We continue this cycle as shown in Figure \ref{fig:dataset} again and again until we get enough high-quality data. After several cycles, a high-quality dataset that meets human requirements is finally obtained.

Here, we choose LoRA for fine-tuning due to its efficient parameter use, low inference overhead, and broad applicability. LoRA strikes a balance between performance and ease of use. It also benefits from strong community support and seamless integration with libraries.

We use the dataset to fine-tune our SLMs so that they will output the correct JSON text as expected. Here we use Huggingface's PEFT library \cite{peft} to help us quickly implement this process. During the fine-tuning process, we adjust parameters such as learning rate and epochs to obtain the accuracy of the SLMs on the test set under different parameters for comparison. The details of fine-tuning are displayed in Section \ref{sec: experiments and results}.

\begin{figure*}[htb]
    \centering
    \includegraphics[width=1.8\columnwidth]{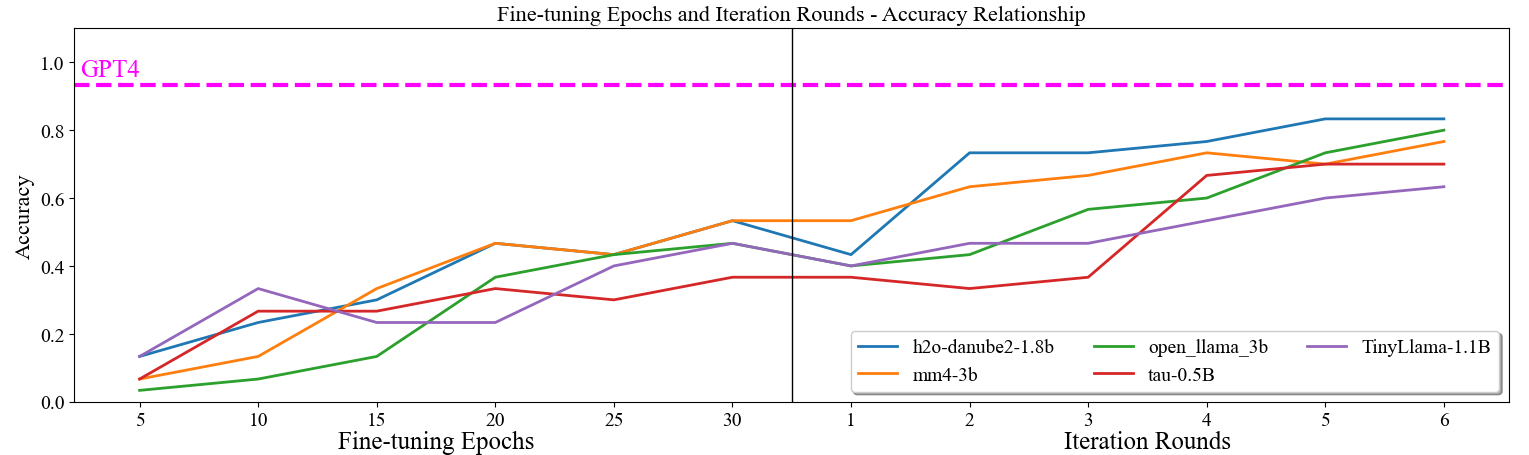}
    \caption{\textbf{The accuracy curve during fine-tuning and iteration.} This graph shows the accuracy of five models and GPT4 on the test set during fine-tuning and iteration processes. It can be seen that all the models' accuracy rises gradually in both processes and finally gets close to that of GPT4. (The last fine-tuning epoch is immediately followed by the first round of iterations, which is why they continue on the same axis.)}
    \label{fig:acc}
\end{figure*}

\subsection{Teacher-student Iteration}
\label{sec:iteration}

The teacher-student model is a paradigm that has been proven effective in the context of language model training and knowledge transfer \cite{Chen2024GrimoireIA}. Our method is rooted in the principles of multi-agent systems, where multiple agents collaborate and learn from each other to achieve common goals \cite{Hong2023MetaGPTMP} \cite{Wu2023AutoGenEN}. 

This method involves a more knowledgeable teacher model guiding a less knowledgeable student model, with the goal of improving the student's performance through iterative learning and feedback. The overall framework of the method is illustrated in Figure \ref{fig:architecture}.

A strong language model, such as GPT-4, is designated as the teacher, while a weaker SLM acts as the student. In each iteration, the teacher serves as a prompt engineer, responsible for generating suitable prompts for the student based on the current task and map environment information. The teacher also receives feedback from the previous iteration's results and adjusts the prompts' content in real-time based on this feedback. This ensures that the student can make correct judgments and attempt to correct any incorrect judgments as much as possible.

The student is typically a fine-tuned SLM designed to receive prompts from the teacher and generate corresponding outputs based on the current task. During each iteration, the student's output is recorded and used to provide feedback to the teacher in the subsequent iteration.

Algorithm \ref{alg:iter} describes the procedures of the teacher-student iteration.

\begin{algorithm}
\caption{Teacher-student Iteration}
\label{alg:iter}
\begin{algorithmic}[1]
\State \textbf{Inputs}: tasks $T$; environment information $E$; names of models $N$; expected goals $G$
\State $M^{teacher}, M^{student} \leftarrow \texttt{Initialize}(N)$ 
\State $iteration = 1$
\While{$ iteration < max_(iter) $}
\For{i=1 \ldots n}
    \State $\boldsymbol f = \texttt{ReadFeedback}(\boldsymbol {M^{teacher}})$
    \State $\boldsymbol {p^t} = \texttt{GeneratePrompt}(\boldsymbol {M^{student}}, \boldsymbol f, \boldsymbol {T_i})$
    \State $\boldsymbol {result} = \texttt{Completion}(\boldsymbol {p^t}, \boldsymbol {T_i}, \boldsymbol E)$
    \State $\boldsymbol {state} = \texttt{Navigation}(\boldsymbol {result})$
    \If{$result[positions] == G_i$}
    \State $success = 1$
    \Else
    \State $success = 0$
    \EndIf
    \State $\texttt{UpdateFeedback}(\boldsymbol {result}, \boldsymbol {success})$
\EndFor
\State $iteration = iteration + 1$
\EndWhile
\end{algorithmic}
\end{algorithm}

 \begin{figure}[!htb]
    \centering
    \includegraphics[width=1\linewidth]{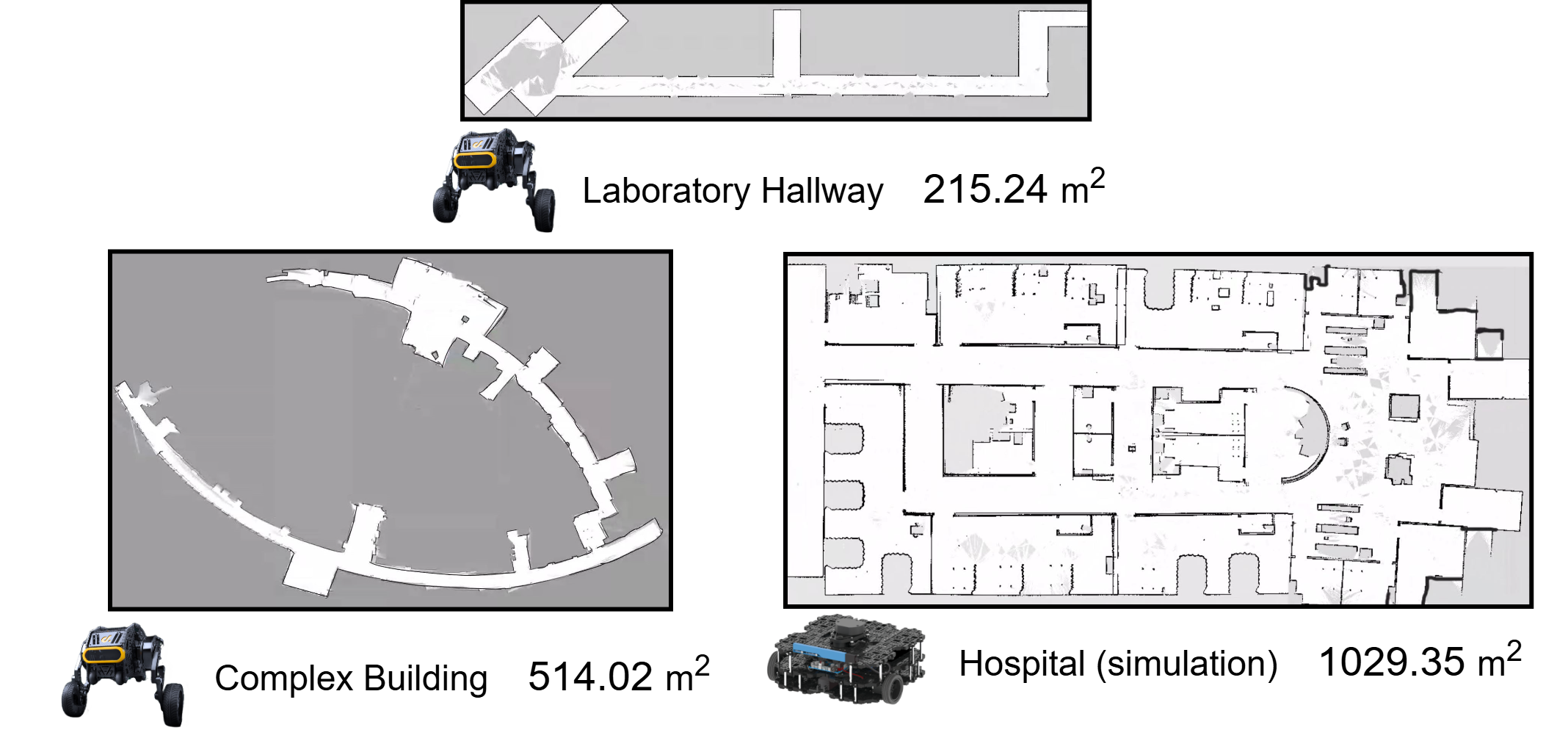}
    \caption{\textbf{The sizes of the environments used in our experiments.} The lab hallway and complex building are used in the real-world experiments, while the hospital is used in the simulation experiment.}\label{fig:size}
\end{figure}

\subsection{Robot Navigation Controller}
\label{sec:navigation}

In our current setup, SLMs serve as task planning tools, taking map landmarks (coordinates and attributes) and task descriptions as input. They generate an ordered sequence of target coordinates, which low-level navigation algorithms use to move the robot. The inputs and outputs of the SLMs can be seen in Figure \ref{fig:f1} and Figure \ref{fig:architecture}. Thus, after gaining the list of goal points from the language model, a robot navigation controller is required to direct the robot toward the points and update the planned path in real-time.

Here we choose Navigation2 \cite{Macenski2020TheM2}, a modular navigation framework. We extract the list of goal points from the SLM's output and put it into Navigation2. The Nav2 controller then helps the robot navigate from the initial position to the expected points sequentially according to the list.

With all of these components, FASTNav is able to finish the entire navigation process. It receives a task command described in natural language, analyses the task logically, and navigates the robot to the right points sequentially.

\section{Experiments and Results}
\label{sec: experiments and results}

\subsection{Setup}
\textbf{Environment and Dataset.} We use a virtual environment for simulation and two real environments for testing the robot. The simulation environment employs a hospital scenario \cite{hospital}, in which we design approximately 1400 tasks. These tasks are distributed with single goal-point tasks, two-goal-point tasks, three-goal-point tasks, and multiple (four or more) goal-point tasks in a ratio of 1:3:2:1. We use the data to fine-tune the model and additionally design about 100 tasks as a test set.

We use the corridor of our laboratory and a complex building as the real environments. Considering the limitations of real robot experiments, we design only about 100 tasks here for fine-tuning, using 10 tasks to test the robot and verify the feasibility of our approach to the robot. Figure \ref{fig:size} shows the sizes of these different environments.

\textbf{Implementation Details.} An NVIDIA GeForce RTX 4090 is used for model fine-tuning and iteration. For simulation experiments, we select a TurtleBot3 Waffle. For robot experiments, we choose Direct Drive Tech's DIABLO, a direct-drive, wheel-footed robot. We deploy our system on the robot using an NVIDIA Jetson Orin NX 16GB as an edge device.

\textbf{Methods of Evaluation.} We evaluate our method and comparison cases in terms of \textit{accuracy, efficiency and robustness}. We choose 4 metrics to evaluate our method: \textit{Success Rate (SR), Navigation Error (NE), Average Time (AT), Moving Time Ratio (MTR)}.

\begin{equation}
\label{SR}
    SR = \frac{1}{n} \sum_{i = 1}^{n} S_{i}
\end{equation}

\begin{equation}
\label{NE}
    NE = \frac{1}{n} \sum_{i = 1}^{n} d(\hat{g_i},g_i)
\end{equation}

\begin{equation}
\label{AT}
    AT = \frac{1}{n} \sum_{i = 1}^{n} T_i
\end{equation}

\begin{equation}
\label{MTR}
    MTR = \frac{1}{n} \sum_{i = 1}^{n} \frac{T_{mi}}{T_i}
\end{equation}

In Equation \ref{SR}, $S_i$ is a flag for each task's success or failure. $S_i$ is 1 if the task succeeds and 0 if it fails. In Equation \ref{NE}, $\hat{g_i}$ is the estimated value of the $i$-th goal point, and $g_i$ is the ground truth of the  $i$-th goal point. $d(\hat{g_i},g_i)$ means the distance between $\hat{g_i}$ and $g_i$. In Equation \ref{AT}, $T_i$ is the time spent finishing each task. As for Equation \ref{MTR}, $T_{mi}$ is the time spent by the robot to move while executing the $i$-th task. We denote MTR by the ratio of $T_{mi}$ to $T_i$.

\textbf{Comparison cases.} Five SLMs and other compression methods are used for comparison. Due to hardware performance and energy consumption constraints, running LLMs with over 10B parameters on edge devices is difficult, so we select SLMs with fewer than 3B parameters. We choose 5 SLMs that perform well on public benchmarks according to the Open LLM Leaderboard on HuggingFace, as shown in Table \ref{tab:model_performance}. Some models obtained through quantization and knowledge distillation were also selected. Quantization method GTPQ \cite{Frantar2022GPTQAP} is used on Llama3-8B \cite{llama3modelcard} and Mistral-7B \cite{Jiang2023Mistral7}, while an int4 version of Qwen-7B-Chat \cite{Bai2023QwenTR} and MiniMA-3B \cite{Zhang2023TowardsTL} that is distilled from LLaMA2-7B are also selected.

To compare FASTNav with other LLM-based planning and navigation methods, we design three comparison setups and compare them with FASTNav in our simulation environment. LLM+P \cite{LLM+P} integrates LLMs with classical planners by converting natural language tasks into Planning Domain Definition Language (PDDL) for task-solving, then translating the solution back into natural language. AutoTAMP \cite{AutoTAMP} uses LLMs to translate tasks into signal temporal logic (STL), which a task and motion planner (TAMP) uses to generate motion plans. LLM-As-Task Planner \cite{AutoTAMP} lets LLMs directly generate subtasks from instructions, with motion planning handled by a separate algorithm.

\begin{table}[htb]
\centering
\caption{Attributes of Different SLMs}
\label{tab:model_performance}
\begin{adjustbox}{width=1\linewidth}
\begin{tabular}{lcccc}
\toprule
\textbf{Model} & \textbf{Params (B)} & \textbf{Performance (\%)} & \textbf{Memory (GB)} & \textbf{Tokens/s} \\
\midrule
tau-0.5B       & 0.5  & 36.68 &  4.2  &  16.90 \\
TinyLlama-1.1B & 1.1  & 36.42 &  6.3  &  12.51 \\
danube-1.8     & 1.83 & 48.72 &  6.7  &  13.80 \\
openllama-3B   & 3    & 38.26 &  10.7 &  8.42  \\
mm4-3B         & 3    & 53.22 &  11.3 &  7.19  \\
\bottomrule
\end{tabular}
\end{adjustbox}
\end{table}

\subsection{Results and discussion}
We start by testing the inference speed and computational resource usage of several lightweight language models on the NVIDIA Jetson Orin NX to evaluate their performance on an edge device. Next, we modify the SLMs with FASTNav. To verify the effectiveness of our method, the SLMs with FASTNav are compared with some quantified and distilled models with similar memory usage. We also test FASTNav and 3 different LLM-based planning and navigation methods in the simulation environment. Through ablation studies, we determine the specific roles of fine-tuning and iteration in FASTNav. At last, FASTNav is used on a real robot to make an instant and accurate local deployment.

\textbf{The inference speed of SLMs tends to increase as the number of model parameters decreases, while memory usage increases proportional to the number of parameters.} The models listed in Table \ref{tab:model_performance} are deployed on the NVIDIA Jetson Orin NX to explore how their inference speed and computational resource requirements vary with different parameter counts. We measure the inference speed of the models in terms of tokens generated per second and the computational resources consumed in terms of memory used at runtime. The results in Table \ref{tab:model_performance} show that inference speeds up for different models as the number of parameters decreases. And models' memory requirements increase with the number of parameters. Overall, the results demonstrate the feasibility of using SLMs for lightweight inference at the edge. We look forward to following up with a model that is both fast and accurate enough to be suitable for local deployment on robots.

 \begin{figure}[htb]
    \centering
    \includegraphics[width=1\linewidth]{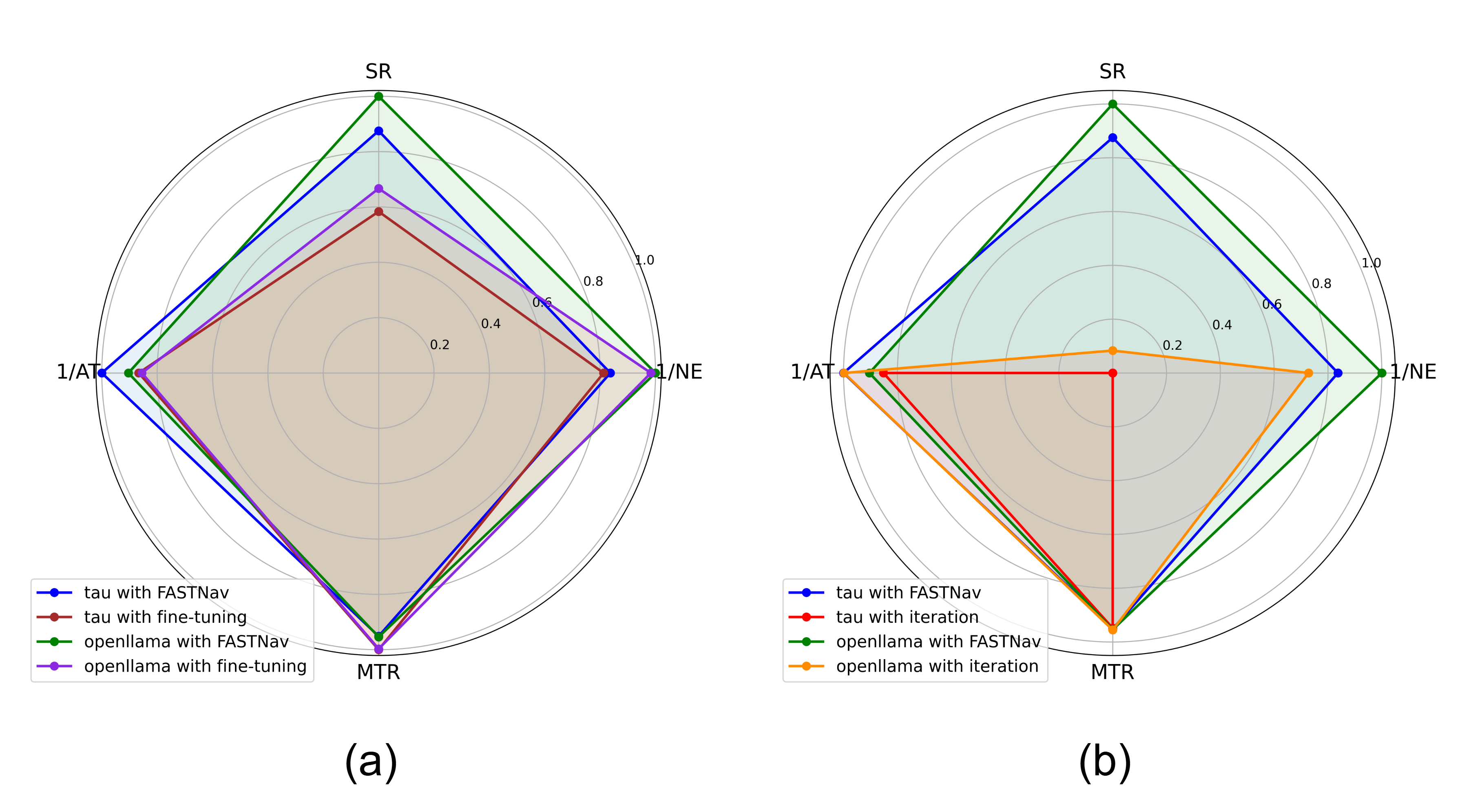}
    \caption{\textbf{Ablation experiments. (a)} FASTNav method and fine-tuning-only method. Through FASTNav, models apparently have better SR than those only fine-tuned. \textbf{(b)} FASTNav method and iteration-only method. SLMs without fine-tuning perform terribly compared to SLMs with FASTNav.}\label{fig:ablation}
\end{figure}

\textbf{SLMs, after FASTNav's fine-tuning and iteration, achieve a 4-fold increase in accuracy while maintaining their lightweight nature.} Through fine-tuning, the model's output format is standardized, making it easier for the robot to read coordinates. Additionally, the model's understanding of maps deepens, demonstrating improved responsiveness to human commands that do not contain specific location information. Multiple rounds of fine-tuning on the training dataset encode the relevant positional information into the model's parameters. We record the model's performance on the test set after each epoch, using accuracy as the metric. To further improve the performance of SLMs, we test our method in the simulation environment. Five selected models iterate through the teacher-student iteration to continuously optimize their performance. We recorded the model's accuracy after each iteration and plotted the curve showing the accuracy changes with the iteration number. Figure \ref{fig:acc} shows that, during fine-tuning, as the number of training epochs increases, the model's accuracy gradually improves and stabilizes around a certain value. The final results far exceed those of the original models, approaching the level of GPT-4, which aligns with our expected experimental results. This conclusion applies generally to the five models we selected, indicating a universal pattern. Therefore, we consider FASTNav to be effective for multi-point navigation tasks.

\begin{figure*}[htb]
    \centering
    \includegraphics[width=1.7\columnwidth]{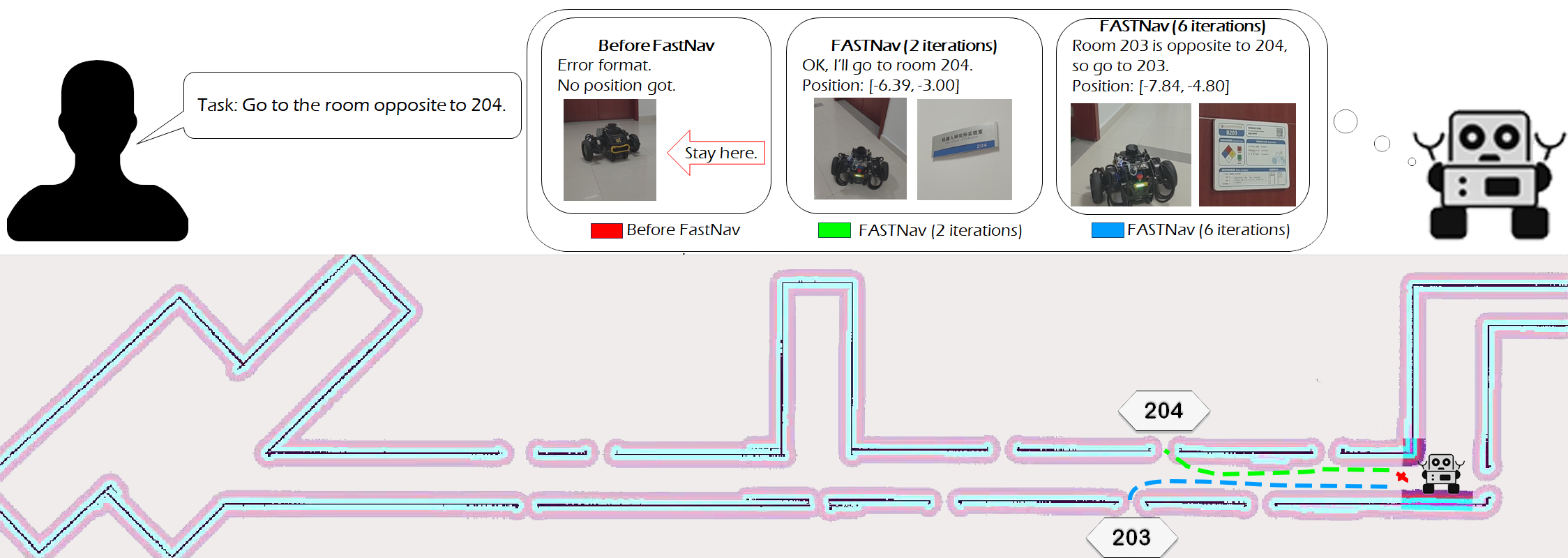}
    \caption{\textbf{Performance of the language model on one task under different conditions.} The image compares the model's performance on one task on a real robot in 3 different configurations: without FASTNav, with FASTNav (2 iterations) and with FASTNav (6 iterations). The robot's capabilities significantly improved before and after FASTNav.}
    \label{fig:tasks}
\end{figure*}

\begin{table}[htb]
\centering
\caption{Comparison of Model Performance between FASTNav and Other Compression Methods}
\label{tab:model_comparison}
\begin{adjustbox}{width=1\linewidth}
\begin{tabular}{lcccc}
\toprule
\textbf{Models} & \textbf{Memory (GB)} & \textbf{NE (m)} & \textbf{SR (\%)} & \textbf{MTR} \\
\midrule
Llama3-8B-GPTQ & 8 & 0.25 & 36.67 & 0.941 \\
Mistral-7B-GPTQ & 6 & 0.32 & 23.33 & 0.943 \\
Qwen-7B-Chat-Int4 & 9 & 0.21 & 13.33 & 0.945 \\
distilled-MiniMA-3B & 8 & 0.16 & 6.67 & 0.966 \\
\textbf{openllama-3B with FASTNav} & 8 & 0.2 & 70 & 0.991 \\
\textbf{tau-0.5B with FASTNav} & 2 & 0.19 & 63.33 & 0.993 \\
\bottomrule
\end{tabular}
\end{adjustbox}
\end{table}

 \begin{figure}[htb]
    \centering
    \includegraphics[width=1\linewidth]{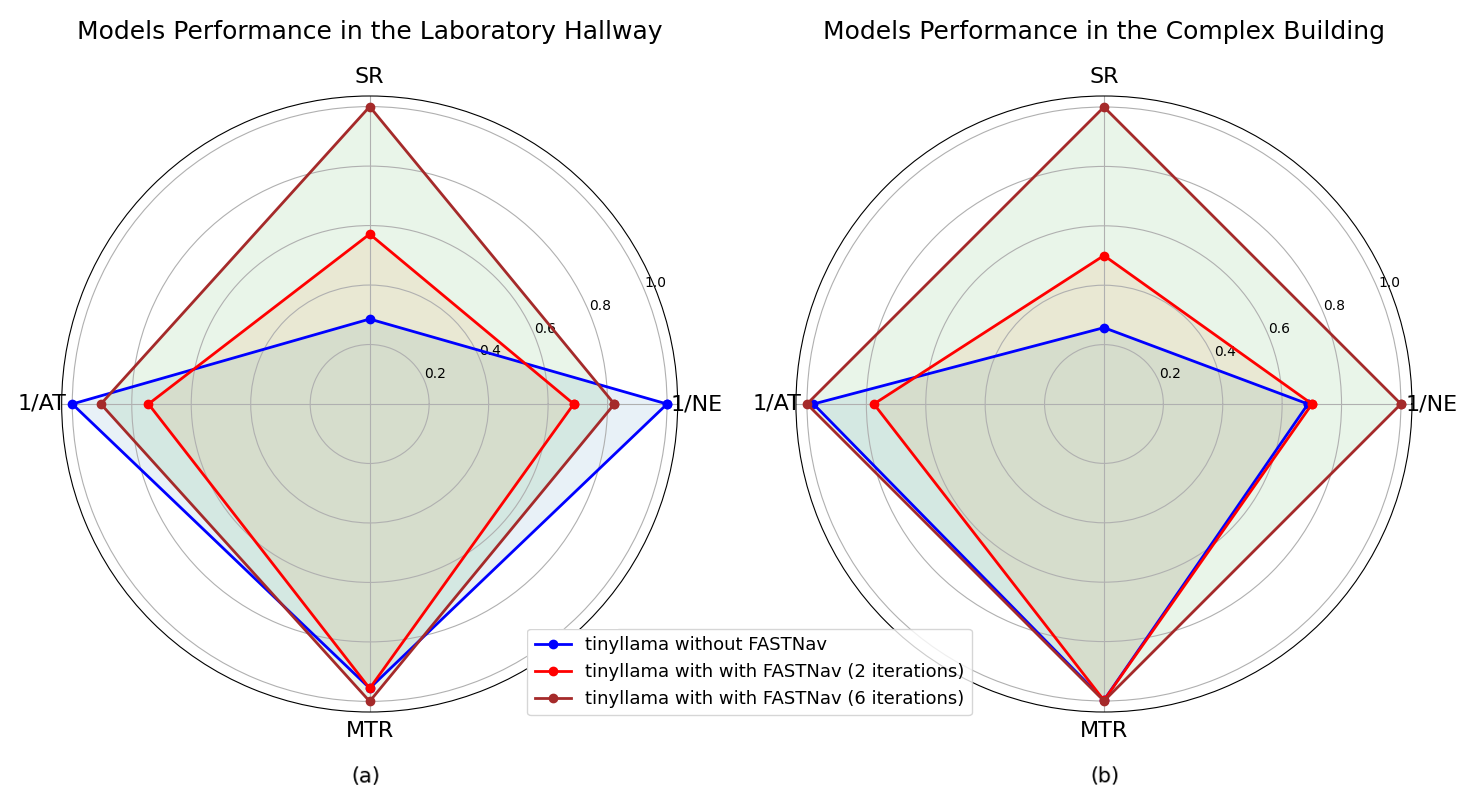}
    \caption{\textbf{Experiments on a real robot. }After being processed by FASTNav, TinyLlama-1.1B shows higher SR and better performance on a real robot than before adjustments. Increasing the number of iterations also enhances the model's capabilities. We conducted experiments in two real-world environments of different sizes, demonstrating the method's real-world viability.}\label{fig:realnav}
\end{figure}

\textbf{Compared to the selected LLM compression methods, FASTNav demonstrates an improvement of at least 30$\%$ in success rate and maintains a quicker response speed for navigation.} Table \ref{tab:model_comparison} shows the memory usage of some comparisons and our chosen SLMs. Most models here cost 7-9 GB, which means the difficulties of deploying them locally are about the same. The SLMs with FASTNav here have the memory during the iterations, and the teacher models are removed so that they can respond fast. It can be seen that FASTNav with similar memory usage is remarkable, having better performance than comparisons. Regarding NE and SR, we can summarize that SLMs with FASTNav are more suitable for specific tasks. This is the result of fine-tuning and iteration. For AT and MTR, it is evident that our method is lighter and faster, resulting in more time spent on robots' movement instead of language models' inference. It may be because after fine-tuning, the models have learned the map information in the environment well, so we do not include this part in the prompt during querying. This allows the models' inputs to be drastically scaled down, with a corresponding reduction in the inference time. Especially for tau-0.5b with FASTNav, we find it has a fascinating performance that is close to that of much larger models, even with only 0.5B parameters. This phenomenon shows that FASTNav is able to fully tap the potential of SLMs, making it possible to deploy language models with low cost but high performance.

\begin{table}[htb]
\centering
\caption{Success Rate and Inference Time Comparison}
\label{tab:baseline}
\begin{adjustbox}{width=1\linewidth}
\begin{tabular}{lcc}
\toprule
\textbf{Method} & \textbf{Success Rate (\%)} & \textbf{Inference Time (s)} \\
\midrule
LLM+P (GPT4) & 76.67 & 15.38 \\
LLM+P (openllama-3B) & 0 & — \\
AutoTAMP (GPT4) & 53.33 & 224.66 \\
AutoTAMP (openllama-3B) & 0 &— \\
LLM-As-Task Planner (GPT4) & 46.67 & 34.24 \\
LLM-As-Task Planner (openllama-3B) & 0 & —\\
\textbf{FASTNav (openllama-3B)} & 70 & 2.87 \\
\bottomrule
\end{tabular}
\end{adjustbox}
\end{table}

\textbf{FASTNav demonstrates clear advantages in both performance and inference speed compared to other LLM-based planning and navigation methods.} Based on the results in Table \ref{tab:baseline}, FASTNav shows clear advantages in robot navigation tasks compared to other methods. Unadjusted SLMs like openllama-3B struggle with tasks that require converting natural language into structured formats such as PDDL or STL, highlighting the importance of fine-tuning to constrain output formats. FASTNav, with fine-tuning and teacher-student iteration, achieves a 70$\%$ success rate, outperforming most methods using GPT-4, except for LLM+P, which has a slightly higher success rate of 76.67$\%$. Additionally, FASTNav’s inference time of 2.87 seconds is significantly lower than the other methods, which require complex reasoning and multiple rounds of prompts. This balance of efficiency and accuracy, along with its competitive performance, indicates that FASTNav is well-suited for real-world applications, with strong potential for future scalability and broader applicability in navigation tasks.

\textbf{Through ablation, we find that fine-tuning can effectively constrain the output format and enhance task success rate, while teacher-student iteration is responsible for further improving the performance of fine-tuned models (reaching near teacher performance).} To verify every module's role in our method, we design ablation experiments. For two chosen models, openllama and tau, we test their performance in the simulation environment in the case of \textit{with FASTNav}, \textit{with iteration only}, and \textit{with fine-tuning only}. Results in Figure \ref{fig:ablation}(a) show that fine-tuning is the key to using SLMs on specific domains, which greatly constrain the output format of SLMs. Without fine-tuning, it is really hard for SLMs to give outputs truly suitable for our tasks. According to Figure \ref{fig:ablation}(b), on the basis of fine-tuning, the teacher-student iteration can further improve their abilities, leading to a 30$\%$ - 40$\%$ SR increment. If we were to compare the performance of the two individually, we found that teacher-student iteration alone is inferior to fine-tuning alone. This is likely because fine-tuning reinforces map information in the SLM and constrains the output format, which is crucial for our later use. In contrast, iteration alone is less effective due to the SLM's limited generalization capabilities. Combining these two modules, FASTNav is proven to go beyond fine-tuning-only or iteration-only methods.

\textbf{FASTNav is an effective method for enhancing SLM performance, creating high-performing and edge-deployable language-guided robot navigation models.} We deploy FASTNav on DIABLO, conducting experiments in a laboratory hallway and a complex building, and the results are shown in Figure \ref{fig:realnav} and Figure \ref{fig:tasks}. We chose the TinyLlama-1.1B model for experiments on the robot. As shown in Figure \ref{fig:realnav}, the unadjusted TinyLlama language model has a low success rate and significant navigation errors. After 2 iterations and 6 iterations, the model's accuracy gradually increases, reaching a high level. The navigation errors gradually decrease, and its usability in real environments steadily improves. Based on a series of data, the language model fully adjusted by FASTNav also demonstrates superior performance in real environments.

\subsection{Limitations}
\begin{itemize}
    \item FASTNav currently focuses on language models, missing other multimodal information like vision.
    \item FASTNav needs a large amount of data for fine-tuning, which is not always easy to get.
    \item FASTNav is primarily task-oriented, requiring specific fine-tuning and iteration for each different task environment.
\end{itemize}

In the future, we aim to explore the lightweight compression of multimodal language models that can process both visual and textual information, allowing our method to address more complex tasks by leveraging richer environmental data. Additionally, we plan to refine the structure of our approach to enhance its few-shot learning capability, reducing its reliance on extensive fine-tuning. This will improve its adaptability to diverse environments and enable it to perform well even with limited task-specific training data.

\section{Conclusion}
\label{sec:conclusion}
In this work, we propose FASTNav, which represents fine-tuned adaptive small language models trained for multi-point robot navigation. This method includes three modules: fine-tuning, teacher-student iteration, and robot navigation controller. It is shown that SLMs can remain lightweight and have great performance close to much larger models in specific domains with FASTNav. We believe that FASTNav can greatly release the potential of SLMs and drive the widespread application of LLM technologies at the edge end.


\bibliographystyle{IEEEtran}
\bibliography{references.bib}

\end{document}